\documentclass{article}

\usepackage{graphicx} 
\usepackage[a4paper, total={6.15in, 9.7in}]{geometry}
\usepackage{amsmath}
\usepackage{amsthm}
\usepackage{amssymb}
\usepackage[english]{babel}
\usepackage{authblk}
\usepackage{booktabs}
\usepackage{hyperref}
\usepackage{subcaption}
\usepackage{orcidlink}
\usepackage{dsfont}
\usepackage{cases}
\usepackage{xcolor} 
\usepackage{bm}
\usepackage{tikz}

\title{Interval Forecasts for Gas Prices in the Face of Structural Breaks -- Statistical Models vs. Neural Networks}

\author[1]{Stephan Schlueter} 

\author[2]{Sven Pappert}

\author[3]{Martin Neumann}

\affil[1]{Institute of Energy Engineering and Energy Economics, Ulm University of Applied Sciences, Prittwitzstrasse 10, 89233 Ulm, Germany, Stephan.schlueter@thu.de,
ORCID: 0000-0001-5816-3337}

\affil[2]{Faculty of Statistics, TU Dortmund University, Vogelpothsweg 78, 44227 Dortmund, Germany, pappert@mail.statistik.tu-dortmund.de, ORCID: 0009-0004-1474-1958}

\affil[3]{86154 Augsburg, Germany}

\begin{document}

\maketitle

\abstract{Reliable gas price forecasts are an essential information for gas and energy traders, for risk managers and also economists. However, ahead of the war in Ukraine Europe began to suffer from substantially increased and volatile gas prices which culminated in the aftermath of the North Stream 1 explosion. This shock changed both trend and volatility structure of the prices and has considerable effects on forecasting models. In this study we investigate whether modern machine learning methods such as neural networks are more resilient against such changes than statistical models such as autoregressive moving average (ARMA) models with conditional heteroskedasticity, or copula-based time series models. Thereby the focus lies on interval forecasting and applying respective evaluation measures. As data, the Front Month prices from the Dutch Title Transfer Facility, currently the predominant European exchange, are used.
We see that, during the shock period, most models underestimate the variance while overestimating the variance in the after-shock period. Furthermore, we recognize that, during the shock, the simpler models, i.e. an ARMA model with conditional heteroskedasticity and the multilayer perceptron (a neural network), perform best with regards to prediction interval coverage. Interestingly, the widely-used long-short term neural network is outperformed by its competitors.} \\




\section{Introduction}
Until mid 2021 the European gas markets were characterized by oversupply and constant increase of liquidity. The North Stream 2 pipeline was about to be opened up and its predecessor North Stream 1 reliably supplied Europe with a significant amount of Russian gas. Comparably simple autoregressive gas price models were sufficient to capture the dynamics of short-term gas prices. In the prelude to the war in Ukraine Gazprom, i.e. the company supplying the gas for the North Stream 1 pipeline, significantly reduced the gas flows towards Europe starting a period of price turmoil including spikes and high volatility. Due to a warm Winter 2023/23 and resulting oversupply the markets calmed down. However, for risk management and market analysis the problem of at least one structural break remains and it needs to be verified whether existing stochastic price models still can be used. Especially the calibration process of stochastic models who rely on historic prices is an issue. 

In this article we analyze the effects of the aforementioned geopolitical changes on short-term risk management by comparing the performance of different models. Here, probabilistic, especially interval forecasts are of interest instead of mean forecasts, which are commonly considered. Thereby we use prices from the Dutch Title Transfer Facility (TTF), currently the most liquid virtual trading hub in Europe, as an example. Thereby we assume that the efficient market hypothesis holds, although liquidity levels especially in the forth quarter of 2022 raise some doubts about it. Regarding models we especially focus on the variance as it shows that the temporal dependence of short term commodity future price time series is dominated by the second moment. As fundamental benchmark for all tested models the widely-used autoregressive moving average model (ARMA) \cite{hamilton2020time} is used, but with conditional heteroskedasticity. Thereby a version of the GARCH model (precisely, the asymmetric power GARCH) is applied, whereby GARCH stands for generalized autoregressive conditional heteroskedasticity \cite{bollerslev1986generalized, embrechts2011quantitative}. The model is an autoregressive model for conditional volatility especially designed to capture volatility clusters. Eventually, as suggested by \cite{loaiza2018time} and \cite{pappert2023forecasting}, based on the works of \cite{chen2006estimation, beare2010copulas} we also test a copula-based time series approach. Calibration of the latter model is not easy and theoretic requirements are high (e.g. stationarity) in order to guarantee stable/reliable estimates. This is why we choose artificial neural networks (ANNs) as a potential alternative. These machine learning models have proven powerful in multiple applications \cite{hill1996, hossain2020}. We consider various types of neural networks, whereby we especially consider recurrent neural networks such as temporal convolutional networks or the widely-used long-short term (LSTM) model that both allow incorporating temporal dependencies. All models are evaluated given a number of metrics such as the average coverage, width, the interval score and pinball (PB) losses \cite{gneiting2007strictly}. The resulting values allow to quantify the predictive performance of a certain method. Besides, we also analyse the performance graphically.

Results show that the simplest model, i.e. an autoregressive model with time-varying conditional volatility and the multilayer perceptron (MLP), outperform all other competitors during the shock. Thereby it is remarkable that the LSTM model, which is widely used in time series forecasting, is clearly outperformed by the very simple multilayer perceptron network regarding interval precision -- presumably due to very small data set (at least in machine learning standards) \cite{ormaniec2022}. To be more precise: In the period before the price increase in Summer 2022, all models and also the neural network-based ones perform reasonably well. However, especially during the shock period between Summer 2022 and Spring 2023, it shows that simplicity rules. In general interval width of the neural network-based models is too small in the shock period and too large in the post-shock period. 

The paper is structured as follows: In Section \ref{predInt} we introduce prediction intervals, evaluation measures, and prediction methods, whereby in Section \ref{secModels} the tested models are presented. In Section \ref{secCaseStudy}, the predictive performance of all considered models is evaluated. The data set is presented and results are discussed thoroughly. Section \ref{secConclusion} concludes the article.

\section{Prediction Intervals: Definition, Methods, and Goodness-of-fit Measures}
\label{predInt}

Predicting intervals differ from point forecasts in certain ways: First, additional information is needed regarding interval width (e.g. 95\% interval or 50\% interval), whereby wider intervals correspond to higher uncertainty in the prediction. Second, quality measurement is less straightforward than when measuring point forecasts (e.g. root mean squared error), and we discuss this matter in detail in Section \ref{secGoF}: Wider intervals are more likely to cover the realized values but contain less information. The goal is to produce intervals as narrow as possible while still ensuring the chosen confidence level.

Now let $\{Y_t\}_{t \in \mathbb{Z}}$ be a time series. Let a prediction interval for $Y_{t+1}$ based on past values be given by an lower and a upper bound, $L_{t+1}$ and $U_{t+1}$ respectively. We denote a prediction interval as the $(1 - \alpha)$ prediction interval if $\mathbb{P}(Y_{t+1} \in [L_{t+1}, U_{t+1}]) \geq 1 - \alpha$ holds. The width of the interval is given as the difference of the upper and lower bound. 
Given a $(1 - \alpha)$ prediction interval, several evaluation measure can be relevant. We first introduce the relevant evaluation measures in Section \ref{secGoF} and proceed to explain quantile regression (Section \ref{sec:Quantile}) on which the training of the neural network models is based on. 

\subsection{Metrics to Evaluate Prediction Intervals}
\label{secGoF}

The forecasting performance of an $(1-\alpha)$ prediction interval (PI) is essentially assessed by two measures: The PI coverage percentage (PICP) and the PI average width (PIAW). The PICP gives the percentage of how often the actual value lies in the respective PI. Evaluating $n$ PIs on $n$ realizations $y_i, i \in \{1,\hdots,n\}$, the PICP is given by
\begin{align*}
PICP = \frac{1}{n} \sum_{i=1}^{n} c_i, \hspace{10pt} c_i = 
\begin{cases}
          1, & y_i \in [L_i,U_i] \\
          0, & y_i \notin [L_i,U_i] \\
        \end{cases}.
\end{align*}
The PIAW averages the width of each PI, is given by
\begin{align*}
    PIAW = \frac{1}{n} \sum_{i=1}^{n}(U_i - L_i).
\end{align*}
Finding the best prediction interval is finding the best compromise between these two measures.
To evaluate interval predictions with respect to their coverage rate, as well as their width the interval score $S$ from \cite{gneiting2007strictly} may be utilized. It is given by
\begin{align}
    S(U_i,L_i;y_i) = (U_i - L_i) + \frac{2}{\alpha}(L_i - y_i) 1_{y_i < L_i} + \frac{2}{\alpha}(y_i - U_i) 1_{y_i > U_i}.
\end{align}
Here $u$ denotes the upper boundary of the $(1-\alpha)$ PI while $l$ denotes the lower boundary. The formula is valid if the lower boundary is taken as the $\frac{\alpha}{2}$ quantile and the upper boundary as the $1-\frac{\alpha}{2}$ quantile. 

Eventually, to evaluate individual quantile forecasts (i.e. the lower or upper bound of the PI) The pinball loss can be utilized. Note that the pinball loss is also used as loss function for training the NN models. The pinball loss for a certain quantile of a PI is given by 


\begin{align}
\label{eq:pb}
    L_{\alpha}(\hat{q}_{\alpha}, y_i) = 
        \begin{cases}
          (1 - \alpha)(\hat{q}_{\alpha} - y_i), & \hat{q}_{\alpha} \geq y_i \\
          \alpha(y_i - \hat{q}_{\alpha}), & \hat{q}_{\alpha} < y_i. \\
        \end{cases}       
\end{align}

Here $y_i$ again denotes the realized value, $\alpha$ the level of the quantile, and $\hat{q}_{\alpha}$ the predicted quantile.

\subsection{Quantile Regression}
\label{sec:Quantile}

In case of heteroscedastic data methods that construct fixed-length PIs are not suitable as they would assign the same uncertainty to all time steps and hence produce overly conservative (wide) intervals. So, the width of the PI should adapt to the variability in each time step. This is the goal of quantile regression (QR), originally introduced by \cite{koenker1978}. According to  \cite{davino2014} this method tries to estimate a conditional quantile function (CQF) of $Y$ given $X$ at the specified significance level $\alpha, 0 \leq \alpha \leq 1$. The CQF is defined as:
\begin{align*}
    q_{\alpha}(x) = \inf \{y \in \mathbb{R} : F_Y(y|X=x) \geq \alpha \}
\end{align*}
The confidence level $(1-\alpha)$ of the PI is the difference between two quantile levels, so the estimated PI becomes:
\begin{align*}
    \hat{I}_{\alpha}(x) = [\hat{q}_{\alpha_{l}}(x), \hat{q}_{\alpha_{u}}(x)]    
\end{align*}
whereby  $\hat{q}_{\alpha_{l}}$, $\hat{q}_{\alpha_{u}}$ are the derived CQFs for $\alpha_{l} = \alpha / 2$, $\alpha_{u} = 1 - \alpha / 2$. Note that the interval width depends on each $x$ individually. When the ideal interval $I_{\alpha}(x)$ is replaced by the finite approximation $\hat{I}_{\alpha}(x)$ the actual coverage of the PI is not guaranteed to match the confidence level $(1 - \alpha)$.\\
The estimation of $\hat{q}_{\alpha_{l}}$, $\hat{q}_{\alpha_{u}}$ can be turned into an optimization problem with the objective to minimize the pinball loss (see Eq. \ref{eq:pb}):



\subsection{Quality Driven Prediction Interval Estimation}

A modified version of the Lower Upper Bound Estimation (LUBE) \cite{khosravi2011} is the quality driven (QD) loss \cite{pmlr-v80-pearce18a}. Like LUBE, which directly tries to minimize the $PIAW$ while ensuring the $PICP$, the QD method also follows this idea: The training of the network is oriented towards the so called high-quality principle for prediction intervals. But unlike LUBE, the resulting objective function is still compatible with the commonly used gradient-based optimization algorithms. In addition to the originally width and coverage requirements, Pearce et al. also incorporated the sample size $n$ and the confidence level $\alpha$ and designed the loss function showed in Eq. (\ref{eq:qd}).

\begin{align}
\label{eq:qd}
    QD = PIAW_{capt.} + \lambda \frac{n}{\alpha(1-\alpha)}max(0,(1-\alpha) - PICP)^2.
\end{align}

The first part of the sum refers to the interval width. The captured prediction interval average width $PIAW_{capt.}$ denotes the mean width only of the intervals covering the target value:
\begin{align*}
    PIAW_{capt.} = \frac{1}{c} \sum \limits_{i=1}^{n} (U_i - L_i) k_i, 
\end{align*}
with the number of captured data points
\begin{equation*}
    c = \sum \limits_{i=1}^n k_i
\end{equation*}
and the indicator function
\begin{equation*}
    k_i = \mathds{1}_{L_i \leq y_i \leq U_i}.
\end{equation*}
Only accounting for intervals that capture their target point is motivated by the thought that intervals which are missing their target value shouldn't be allowed to shrink any further. The second factor in Eq. (\ref{eq:qd}) considers the actual interval coverage: If it is equal to or better than the target confidence level, nothing gets added and the loss only consists of the captured width. Otherwise a penalty will be applied depending on the deviation from the confidence level, the confidence level itself, the total number of data points and the  Lagrangian parameter $\lambda$. This parameter allows to control the weight of the coverage's influence in relation to the width \cite{salem2020}.

\section{Models}
\label{secModels}

In this section, we describe three models for interval forecasting that are used as competitors in the case study hereafter. Thereby, only a brief introduction is given with respective references for further reading. By the nature of second moment dependencies, the models that are used to model them are non-linear in different senses. While the ARMA-GARCH model captures the second order dependence explicitly, copula-based time series capture the temporal dependence by allowing a heavy-tailed temporal dependence. ANNs model the dependence with much more complexity. We start by describing the ARIMA-GARCH model in Section ~\ref{secModelsARIMAGARCH} and proceed to describe the copula-based time series model in Section ~\ref{secModelsCopula}. ANNs will be explained in Section ~\ref{secModelsANN}.

\subsection{The (Integrated) Autoregressive Moving Average Model with Heteroskedastic Conditional Volatility}
\label{secModelsARIMAGARCH}
Autoregressive effects are apparent in many time series. The most-widely used standard model for modeling autoregressive effects in the mean is the autoregressive moving average model \cite{hamilton2020time, box2015time}. It is especially applied for simulation and analytical purposes, but can also be used for forecasting. The fundamental model equation reads as follows: Let $\{X_t\}_{t \in \mathbb{Z}}$ be a time series, then
\begin{eqnarray}
\label{eq:ARMA}
    X_t = \mu + \sum_{i=1}^{p} \phi_i X_{t - i} + \sum_{i = 1}^q \theta_{i} \varepsilon_{t - i} + \varepsilon_t
\end{eqnarray}
is an ARMA($p$,$q$) process, where $p,q \in \mathbb{N}$ are the model orders, $\varepsilon_t$ is white noise and the parameters are given by $(\mu, \phi_1,\hdots, \phi_p,\theta_1,\hdots,\theta_q)$. The AR-parameters $\{\phi_i\}_{i=1}^p$ are chosen in such a way that the roots of the associated characteristic polynomial lie outside the complex unit circle to ensure stationarity \cite{hamilton2020time}. In most cases both $p$ and $q$ are small integers and $\varepsilon_t \stackrel{w.n.}{\sim} \mathcal{N}(0,\sigma^2)$ given a standard deviation $\sigma >0$ \cite{hamilton2020time}. The moving average part is represented by the second sum and accounts for smoothness of the mean. Note that there are various extensions to reflect instationarities, seasonal structures or the dependence on external factors. A widely know version is the ARIMA model, where the 'I' denotes Integration of order $d$, meaning that $d$-differences instead of $X_t$ itself is stationary and modeled by the ARMA-model \cite{hamilton2020time}. A key property of the ARMA($p,q$)-model is that the MA-part allows for a representation of $X_t$ in terms of all its past values. This means that the current random variable $X_t$ is influenced by all past observations with an asymptotically decreasing effect. This feature allows for more persistency in the mean and changes the reaction of the model to shocks in comparison to e.g. AR($p$) models. This is relevant when the model is applied to time series with shocks or structural breaks.

If conditional heteroskedasticity is apparent in the time series, a GARCH model may be considered for modeling the conditional variance. Thereby GARCH stands for generalized autoregressive conditional heteroskedasticity. It describes a discrete-time model where the conditional volatility $\sigma_t^2$ is modeled as an autoregressive variable based on previous values of both, past conditional volatilities and the squared error term \cite{embrechts2011quantitative, francq2019garch, bollerslev1986generalized}. If the conditional volatility is time-varying then $\epsilon_t$ from Eq. \ref{eq:ARMA} is replaced by $ \epsilon_t = \sigma_t \gamma_t$, with $\gamma_t$ being white noise. The corresponding GARCH($1$,$1$) model then reads as follows:
\begin{eqnarray}
\label{eq:GARCH}
\sigma_t^2 = \alpha_0 + \alpha_1 \sigma_{t-1}^2  + \alpha_2 \epsilon_{t-1}^2, 
\end{eqnarray}
whereby $\alpha_0, \alpha_1, \alpha_2>0$. If required, both lag orders in $\sigma_t^2$ can be increased to $r \in \mathbb{N}$ and $s \in \mathbb{N}$. We then speak of a GARCH($r$,$s$) model. For calibration issues and further discussions please refer to \cite{embrechts2011quantitative, francq2019garch}. There are various extensions and modifications of the GARCH approach. Here, due to the specific properties of the data set, we apply the asymmetric power GARCH (short APARCH) model where the power parameter is set to 1. In the APARCH model, the absolute value of the conditional volatility is modeled instead of the squared values. the model was shown to be successful for natural gas forecasting in \cite{berrisch2022distributional}. The model equation, Eq. \ref{eq:GARCH} is modified to
\begin{eqnarray}
\label{eq:APARCH}
\sigma_t = \alpha_0 + \alpha_1 \sigma_{t-1}  + \alpha_2 |\epsilon_{t-1}|.
\end{eqnarray}
Combining the ARIMA and APARCH models, the ARIMA-APARCH model is obtained. The model parameters are estimated by maximum-likelihood estimation. The likelihood can be calculated as the product of conditional densities, whereby  optimization is performed numerically. 

\subsection{Copula-Based Time Series Models}
\label{secModelsCopula}
In this section, we briefly explain the copula-based time series model with a focus on forecasting. For a detailed introduction to copulas, we refer to \cite{joe2014dependence} and for a short introduction to \cite{mcneil2015quantitative}. For an overview of copulas in time series analysis we refer to \cite{patton2012review} or \cite{patton2013copula}. Note that there is an ambiguity to the term 'copula-based time series models' as there are two versions to model time series with copulas. In the first approach, the joint conditional distribution of a multivariate time series is modeled, as in \cite{jondeau2006copula} or \cite{hu2006dependence} for financial time series and in e.g. \cite{aloui2014dependence} or \cite{berrisch2022distributional} for energy time series. The model is also referred to as 'Copula-GARCH'. The second approach is to model the temporal dependence of a univariate time series with a copula as in \cite{chen2006estimation} and \cite{beare2010copulas}. In this paper, we focus on the latter version. Building upon the second approach it is also possible to model the temporal as well as the cross-sectional dependence of a time series jointly with a copula \cite{beare2015vine, smith2010modeling, brechmann2015copar}. For detailed introductions to copula-based temporal dependence models of a univariate time series  we refer to \cite{chen2006estimation} and \cite{pappert2023forecasting}: In this case, the joint distribution of consecutive time series observations is modeled by a copula and marginal distribution. To this end, consider the following decomposition of a univariate and stationary time series, $\{X_t\}_{t \in \mathbb{Z}}$, implied by Sklar's theorem \cite{sklar1959fonctions, joe2014dependence},
\begin{align}
    F_{X_t, X_{t-1}, \hdots, X_{t-p}}(x_t,\hdots,x_{t-p}) = C[F_X(x_{t}), \hdots,F_X(x_{t-p})].
    \label{Sklar}
\end{align}
Here $C$ is the copula of $(X_t,\hdots,X_{t-p})$ i.e. $C$ is the joint distribution of the probability integral transformed (PIT) random variables $(F_X(X_t), \hdots, F_X(X_{t-p}))$.\footnote{The probability integral transformation (PIT) ensures that the transformed random variables are standard uniformly distributed \cite{angus1994probability}.} Note that the notation used above implies strong stationarity i.e. $F_{X_t} = \hdots = F_{X_{t-p}} =: F_X$, where $F_X$ is the unconditional distribution of a time series observation. If the time series is a Markov chain of order $p$, then the decomposition in Eq. (\ref{Sklar}) determines the distributional properties of the time series. One major advantage of this model is that it is able to model marginal properties independently from temporal dependence properties \cite{chen2006estimation, beare2010copulas, beare2015vine, brechmann2015copar}. Furthermore, the model is able to capture non-linear temporal dependence structure by choosing an appropriate copula. For instance, if the time series is volatile, i.e. the dependence structure is heavy-tailed, the t-copula can capture the dependence \cite{pappert2023forecasting, loaiza2018time}. If the time series is more persistent in some regions than in others, the temporal dependence structure is asymmetric and can be captured by the Gumbel or Clayton copula. Since the time series analyzed in this paper is dominated by a volatile dependence structure, the t-copula will be used for modeling the temporal dependence. The t-copula is given by $C[u_1, \hdots, u_d] = t_{\nu, \Sigma}(t^{-1}_{\nu}(u_1), \hdots, t^{-1}_{\nu}(u_d))$, where $t_{\nu, \Sigma}$ is the multivariate distribution function of the t-distribution with degree of freedom parameter $\nu \in (2, \infty)$ and Correlation matrix $\Sigma \in (-1, 1)^d$. Thereby $t^{-1}_{\nu}$ denotes the quantile function of the univariate Student t distribution with degree of freedom $\nu$. For a detailed exposition of the t-copula and related copulas, we refer to \cite{demarta2005t}. 

Depending on the marginal distribution specification, there are different maximum likelihood estimation (MLE) procedures. For this study we are specifying the marginal distribution non-parametrically first using the empirical distribution function and secondly using a kernel-based approach. In this case, the observations can be PIT-transformed and the estimation is the usual MLE on the PIT values \cite{chen2006estimation}.

The probabilistic forecast from a copula-based time series model can be accessed by simulation. To this end, the following procedure is employed \cite{simard2015forecasting}. 1) simulate $N$ i.i.d. samples from the standard uniform distribution, $U_i \stackrel{iid}{\sim} U[0,1], i \in \{1,\hdots,N\}$. 2) transform each sample by $C^{-1}[\cdot | F_X(x_{t-1}),\hdots,F_X(x_{t-p})]$ yielding $V_i = C^{-1}[U_i | F_X(x_{t-1}),\hdots,F_X(x_{t-p})], i \in \{1,\hdots,N\}$. The applied transformation is the quantile transformation where the quantile function of the conditional copula $C[\cdot | F_X(x_{t-1}),\hdots,F_X(x_{t-p})]$ is chosen as transforming function. This ensures that $P(V_i \leq t) = F_{V_i}(t) = C[t | F_X(x_{t-1}),\hdots,F_X(x_{t-p})]$, meaning that the dependence structure between $V_i$ and $F_X(x_{t-1}),\hdots,F_X(x_{t-p})$ is given by $C$. Now we still have to account for the marginal properties. The quantile transformation will be utilized again. 3) Transform each $V_i$ by the quantile function of the marginal distribution, $F_X^{-1}$, yielding $\hat{X}_i = F_X^{-1}(V_i)$. Using this procedure, the probabilistic forecast can be approximated.

\subsection{Artificial Neural Networks}
\label{secModelsANN}

Artificial neural networks are universal function approximators \cite{cybenko1989, hornik1989} and a subset of artificial intelligence models in general. They are inspired by neuroscience as their structure mimics the connection of the neurons of the human brain, which enables them to tackle complex problems such as pattern recognition or prediction. To do so, they are trained on previously seen pairs of inputs and their corresponding labels (i.e. output) $(\mathbf{x}_i,y_i)$ which form the so called training set. Through minimizing a loss function on the training set, the network defines a mapping $f(\mathbf{x},\theta) : \mathbb{R}^p \rightarrow \mathbb{R}$ with $\theta$ being the learned parameters or weights. The minimization is observed on the last part of the training data, the validation set. The training is stopped when the validation loss doesn't shrink any further in a given number of epochs. Subsequently the performance of the model is evaluated through predicting the previously unseen part of the data set, the test data.

In the univariate autoregressive case each input $x_i \in \mathbb{R}^p$ may be a vector consisting of $p$ previous time lags of the target variable $y$. Very similar to the fundamental ARMA equation (\ref{eq:ARMA}) we try to find the right parameters $\phi_i$ but without accounting for the previous errors explicitly. Instead we try to capture the error behavior implicitly by using an adequate amount of weights and by introducing nonlinearities through the application of special transfer functions for the neurons inside the network. Below we briefly introduce three different ANN architectures, which were used in this paper. To describe the basic functionality we first start with the simplest neural network implementation. If not stated otherwise, the following section is based on \cite{goodfellow2016} and \cite{aggarwal2018}.

\subsubsection{Multilayer Perceptron}
\begin{figure}[htb]
	\centering
		\includegraphics[width=0.5\columnwidth]{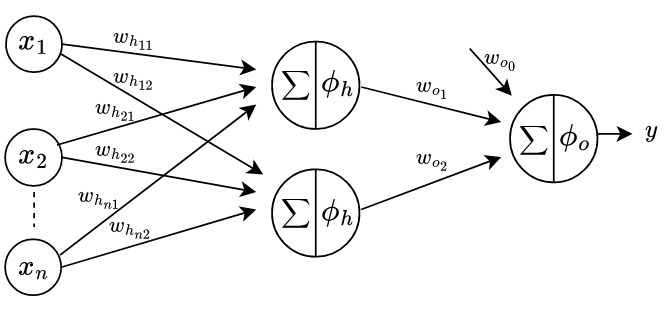}
	\caption{Multilayer perceptron with two hidden neurons and one output. The $n$ features $x_i, i \in \{1,\hdots,n\}$ are first aggregated for each of the two neurons using the weights $w_{hi1}, i \in \{1, \hdots, n\}$ for the first node and the weights $w_{hi2}, i \in \{1, \hdots, n\}$ for the second node. Then the respective aggregate is plugged into the activation function $\phi_h$. The two results are again aggregated using the weights $w_{o_1}$ and $w_{o_2}$. Weighted with $w_{o_0}$ an intercept is added to the aggregate. Lastly, the result is again transformed by an activation function $\phi_o$ to yield the result $y$.}
	\label{fig:nn_ar}
\end{figure}

 MLP's are the most basic type of neural networks. They consist of an input layer, one or multiple hidden layers, and an output layer. Because of the path the information takes straight from the input layer over the hidden layer(s) to the output layer they are also designated as feedforward networks. There are no connections that feed output data back into the model, as opposed to a class of more sophisticated models called recurrent neural networks (RNN) to which the later introduced LSTM network belongs to. To further increase accuracy and also to introduce nonlinearities, one or more hidden layers are stacked between the input and output like seen in Figure \ref{fig:nn_ar}.

Each layer $i$ out of total $n$ layers represents a function $f^{(i)}$ whose output serves as input for the next layer's function i.e. the input passes through multiple chaining functions before it becomes the final output: 
\begin{align}
\label{eq:nn_chain}
    f(x) = f^{(n)}(f^{(n-1)}(f^{(n-2)}(\dots (f^{(1)}(x))\dots))).
\end{align}
The value of $n$ denotes the model's depth, this is why we also talk of ''deep learning''. The width of of each layer is set by the integer $k$ in Figure \ref{fig:nn_ar}(b) which defines the number of neurons or units per layer. Because every hidden layer's neuron is connected to each neuron in the preceding (fully connected) layer, their respective weights are no longer vector- but matrix-valued. To account for the aforementioned nonlinearities, the computed value - the product of input $x$ and weight $W$ plus a constant bias $c$ - in each unit passes an activation function before leaving to the next layer:
\begin{align}
\label{eq:nn_layer}
    h(x) = \phi( W^\top x + c)
\end{align}
The default function recommended for most applications is the rectified linear unit (ReLU) $\phi(z) = max\{0, z\}$ as it still preserves certain linear properties which makes them easy for optimization and good adaptation to unseen data \cite{goodfellow2016}. Figure \ref{fig:nn_ar} shows a simple network that generalizes the AR idea using one hidden layer.

\subsubsection{Temporal Convolutional Networks}
Temporal Convolutional Networks (TCNs) are also a type of feedforward neural network designed specifically for processing sequential data such as time series or audio signals \cite{ashutosh2019}. TCNs leverage the power of convolutional operations, commonly used in image processing, to capture temporal dependencies in the input data. In traditional convolutional neural networks (CNNs), convolutional layers are used to extract local features from spatial data, like images \cite{oshea2015}. TCNs adapt this concept to the temporal domain, where the input is a sequence of data points arranged in a specific order.

The idea behind TCNs is to use one-dimensional convolutions to capture patterns and dependencies across different time steps in the input sequence. The convolutional operation involves sliding a small filter/kernel over the input sequence, computing element-wise multiplications and summing them up to produce a new feature representation, while keeping the original length. The size of the filter determines the extent of the temporal context captured by the convolution operation and also defines how many parameters exist. To avoid large filters while still increasing the temporal coverage of the input sequence (the receptive field of the network) usually multiple convolutions are passed in succession. As effective history increases only linear with the depth of the network, a long input sequence would result in very deep network, that would be hard to train and stabilize \cite{bai2018}.

\begin{figure}[htb]
	\centering
   \includegraphics[width=1.0\columnwidth]{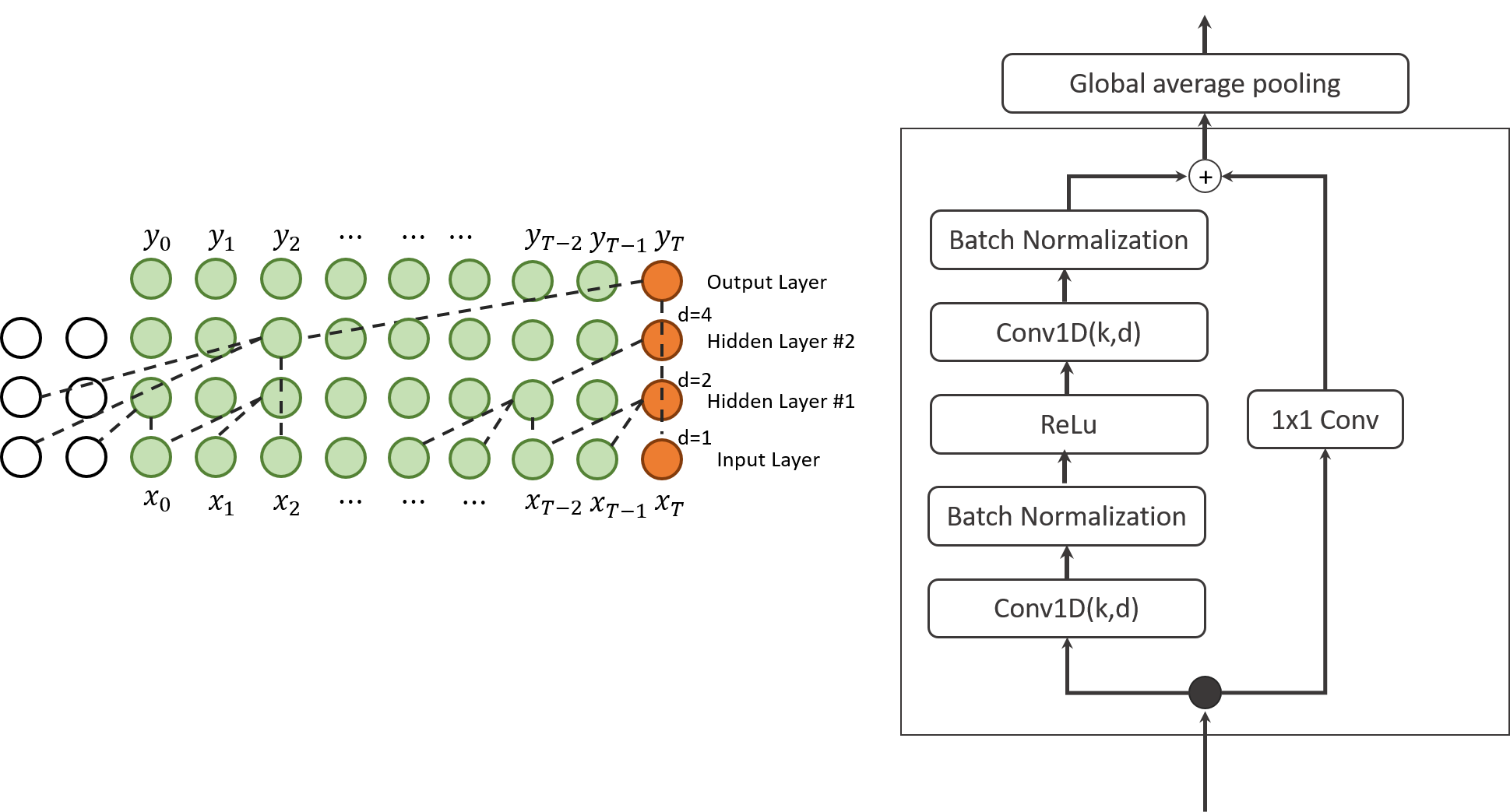}
	\caption{\textit{Left}: A dilated causal convolution with dilation factors d = 1, 2, 4 and filter size k = 3. The receptive field is able to cover all values from the input sequence. \textit{Right}: TCN residual block which may be repeated with different dilations and kernel sizes  \cite{bai2018}.}
	\label{fig:tcn_conv}
\end{figure}

The solution is to apply dilations to the elements of the kernel as showed in Figure \ref{fig:tcn_conv}. With a dilation of $d=2$ for example, the kernel in the second layer now spreads a length of $5$ instead of $3$. Like already stated in the work of \cite{oord2016}, it is ensured that some filter hits each value from the input sequence if the dilation rate increases with a power of $2$. To further provide causality under the constraint that the input and output are of same length, zero padding shall only be applied to the left side of the elements.

To additionally improve performance, two Conv1D layers with the same kernel size and dilation rate are combined and together with normalization and activation layers they form a residual block. \cite{he2016deep} prove that a network based on residual functions is easier to optmize and also counteracts degradation of accuracy with increasing depth. To maximize the effective history captured by the network, multiple instances of this block can be stacked one after the other before the last block is followed by an global average pooling layer to introduce some invariance to small changes in the input.

The fixed filter size which is significantly smaller than the input size results in sparse interactions of neurons. This leads to lower computation time of the weights in contrast to the fully connected MLP or LSTM networks.

\subsubsection{Long-Short-Term-Memory Network}
Long Short-Term Memory (LSTM) networks are a sophisticated type of recurrent neural network (RNN) architecture designed by \cite{hochreiter1997} to overcome typical problems of traditional RNNs (vanishing/exploding gradient) and now belong to the most used RNN models. They have an input and an output layer and one or more hidden layers in between consisting of special LSTM units. These units are more complex compared to the simple perceptrons of the MLP or TCN network. In contrast to the feed forward networks, they preserve some of the information as an internal state, the so called cell state. The preserved information in the cell from time step $t-1$ then gets processed together with the new input in the next time step $t$. To control what part of the information is stored, a LSTM cell has four gates: An input gate, a forget gate, an update gate and an output gate. The first three gates effect the cell state and control what memory content should be replaced \cite{hossain2020}. The output gate determines the degree of exposure of the internal state to the next layers \cite{white2018}. Each gate is triggered by an activation function (sigmoid or tanh) and has its own set of weights and biases which are adjusted during the training process. This ability of memorization enables the LSTM network to handle time dependent relations which makes them particularly well-suited for tasks involving sequential data, such as natural language processing \cite{wang2016}, speech recognition \cite{graves2013}, and time series analysis \cite{sagheer2019}. 

\subsubsection{Technical Aspects of the Neural Network Implementation}

\begin{figure}[htb]
	\centering
		\includegraphics[width=0.4\columnwidth]{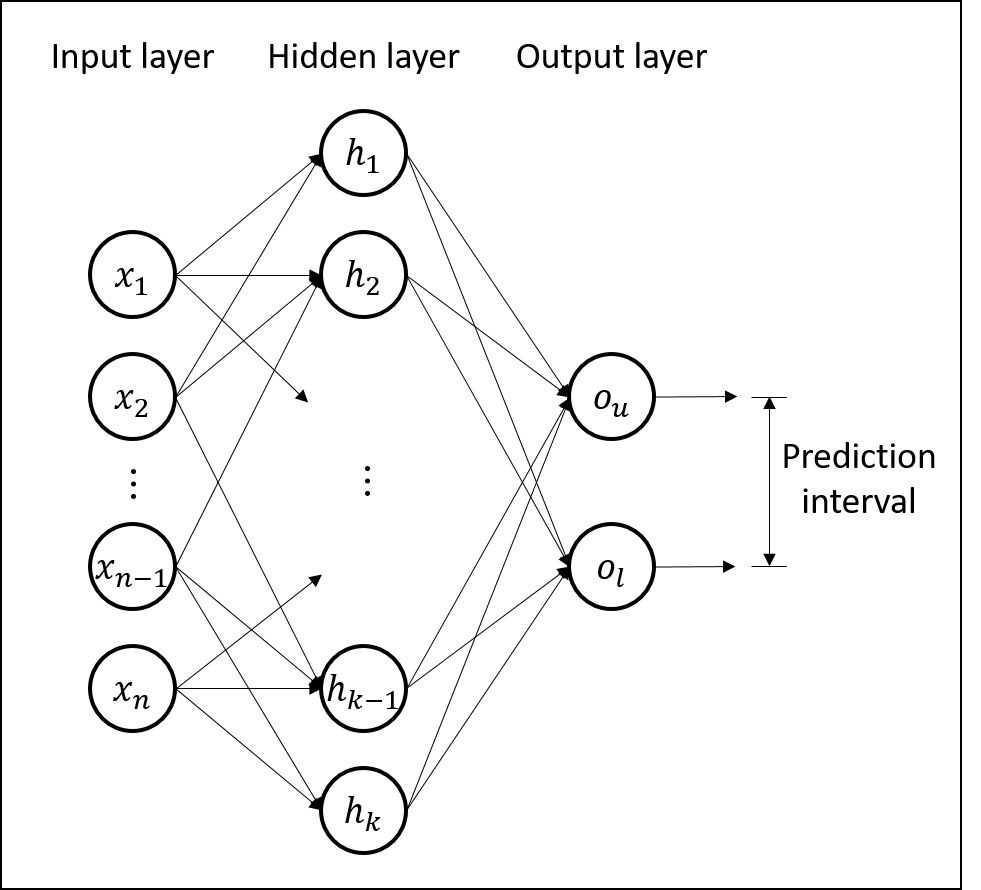}
	\caption{Neural network with one hidden layer and two output nodes for interval prediction \cite{khosravi2011}}
	\label{fig:nn_pi}
\end{figure}

In comparison to point forecasts, the network architectures have to be slightly adapted to interval predictions. Instead of a single output node, we have two nodes which each generate a real value, i.e. one for the lower and one for the upper interval bound, respectively (see Figure \ref{fig:nn_pi}).

Considering the quantile regression approach with the PB loss it would also be possible to add a third output node predicting e.g. the 0.5 quantile or even more nodes/quantiles. This is possible because all quantiles are trained simulatenously \cite{jensen2022}: The average PB loss (Eq. \ref{eq:pb}) over all training points for each quantile $\tau$ is calculated (see Eq. \ref{eq:pbTau}) and then summarized over all $q$ average losses to a combined average loss $PL_c$ as defined in Eq. (\ref{eq:avpl}).
\begin{eqnarray}
    PL_{\tau} = \frac{1}{N} \sum \limits_{i=1}^{N} PL_{\tau}(y_i,\hat{y}_i^{\tau}).
    \label{eq:pbTau}
\end{eqnarray}
\begin{eqnarray}
    PL_c = \frac{1}{q} \sum \limits_{i=1}^{q} PL_{\tau_i}
    \label{eq:avpl}
\end{eqnarray}
This is the final loss the PB-ANN tries to minimize. The advantage is that one can add easily more quantiles, without having to train the network for each quantile individually. Even though including more quantiles it hadn't been found to be a improvement of the overall forecast.

\section{Case Study}
\label{secCaseStudy}

Here the models from Section \ref{secModels} are applied to Dutch natural gas prices, which are described in Section \ref{dataSet}. The tested approaches are fairly complex; hence, in order to see whether this additional modeling and calibration efforts pay off, we benchmark the neural networks and non-linear statistical models with the less complex ARMA-APARCH model. Calibration issues are discussed in Section \ref{subsec:calibration}. For information about neural network training see Section \ref{subsec:NNtraining}. Results are given in Section \ref{Sec:Results}, and a Discussion (Section \ref{subsec:discuss}) concludes the chapter.

\subsection{The Data Set}
\label{dataSet}

\begin{figure}[htb]
	\centering
		\includegraphics[width=1.0\columnwidth]{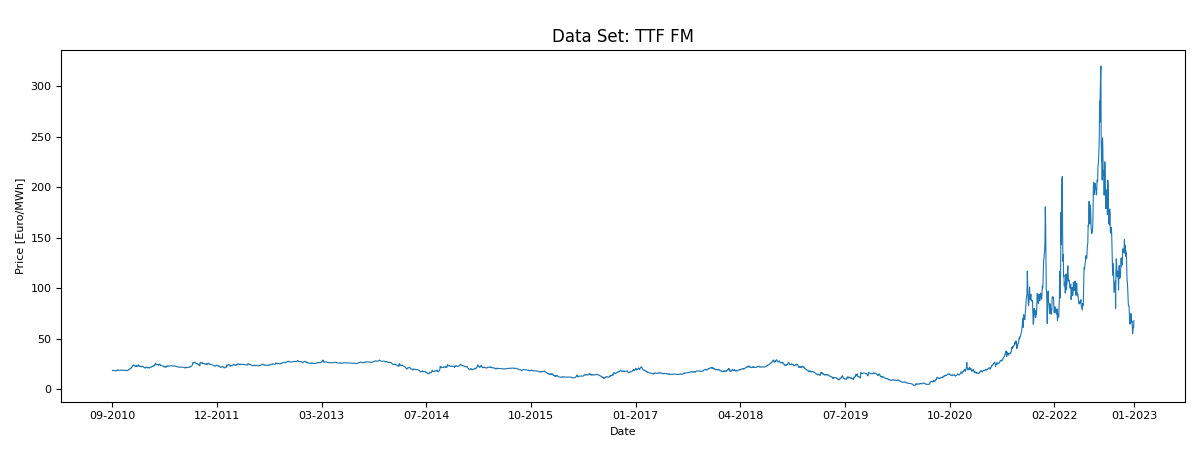}
	\caption{TTF Front Month Prices}
	\label{fig:ttfFM}
\end{figure}

\begin{figure}[htb]
	\centering
		\includegraphics[width=1.0\columnwidth]{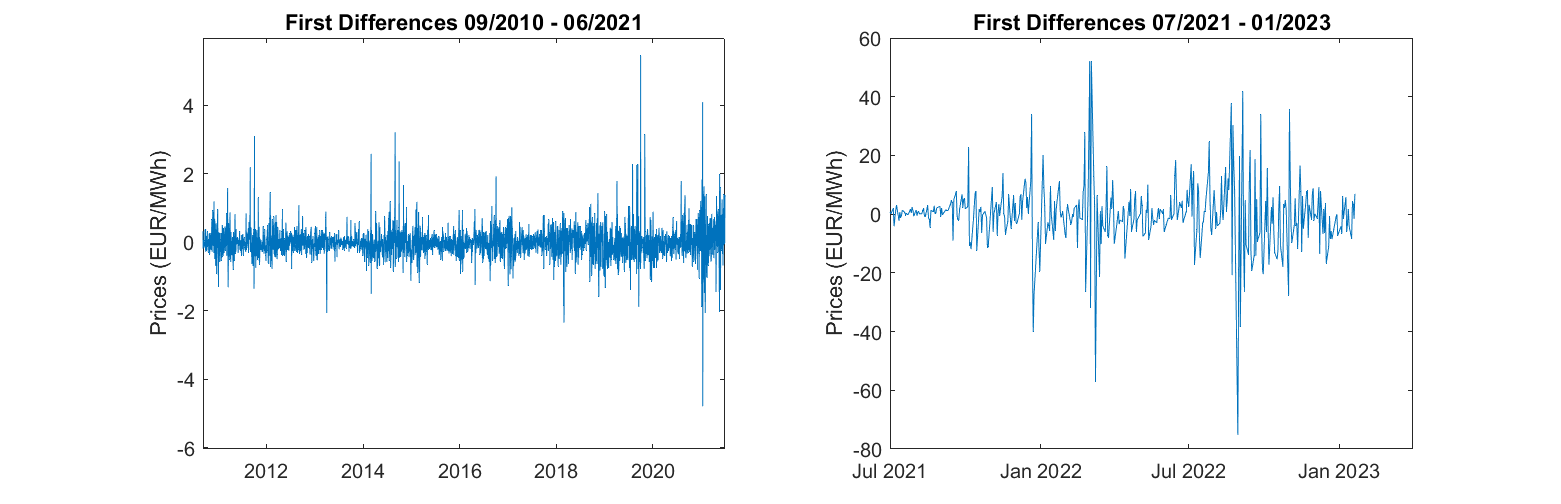}
	\caption{First Differences}
	\label{fig:diffTTF}
\end{figure}

Our data set comprises of 3123 prices for the TTF Front Month (FM)\footnote{Curtesy of WintershallDea AG} starting on 8th of September 2010 until 20th of January 2023. Thereby, trading the FM means trading the right and obligation to receive a constant (physical) supply of 1 MWh of gas in the upcoming month at a certain exit point in the TTF area. Raw data are displayed in Figure \ref{fig:ttfFM}, where we can clearly see the considerably increasing price levels in mid 2022. This is why we additionally show the first price diffferences separated into two periods (before and after 07/2021) in Figure \ref{fig:diffTTF}.

Given this significant structural break, we compute a number of basic descriptive statistics such as empirical variance, skewness, and kurtosis separately for the time before July 2021 and after. Thereby note that we consider first differences. We also perform the Augmented Dickey-Fuller (ADF) test on stationarity of the first moments \cite{embrechts2011quantitative}: if the null hypothesis is rejected (which is the case) the first moments are considered stationary.
Results for common statistics are given in Table \ref{table:DescriptiveStatistics}. In both periods the differences are more or less symmetric as indicated by the skewness. The kurtosis hints that the error distributions are non-Gaussian. It can be observed that the standard deviation is drastically increased after the shock. Also the kurtosis decreases, indicating heavier tails after the shock (see also Figure \ref{fig:diffTTF}). 


\begin{table}[h!]
 	\centering
 	\begin{tabular}{@{}l | l  l  l  l @{}}  
	\toprule
                & Standard Deviation & Skewness & Kurtosis & ADF Test \\ \midrule
First period    &  $0.4531$   &  $1.1451$      &  $22.8306$        & rejected         \\
Second period   &  $11.5400$  &  $-0.4551$     &  $11.7390$        & rejected        \\ 
\bottomrule
 	\end{tabular}
  	\caption{Descriptive Statistics}
   	\label{table:DescriptiveStatistics}
\end{table}

\subsection{Calibration of the Statistical Models}
\label{subsec:calibration}


The ARMA-APARCH model parameters are fitted using maximum-likelihood estimation (MLE), which is explained in detail in  \cite{francq2019garch}. The likelihood has to be calculated in an iterative scheme. In implementation the \texttt{rugarch} package in \texttt{R} is utilized \cite{ghalanos2018package}. The optimization in the \texttt{rugarch}-package is carried out using the \texttt{solnp} optimizer from \cite{ye1988interior}. We choose the lag orders by AIC and choose the skew student-t distribution as innovation distribution for generality. \\
The copula-based time series model parameters are estimated semi-parametrically. The marginal distribution is specified non-parametrically by a kernel-based approach, as specified in \cite{charpentier2007estimation} where a Gaussian kernel is chosen. As copula, the t-copula is chosen. The model is estimated by MLE.

\subsection{Neural Network Training}
\label{subsec:NNtraining}

All networks are implemented using the Python libraries \textit{Tensorflow 2.12.0} and \textit{Keras 2.12.0}. To train the networks the target days to be predicted are subtracted from the end of the set and the remaining data are divided in training and validation set with a 90/10 ratio. Reshuffling the data to perform cross-validation is not an option here because it violates the sequential nature of the data. Furthermore, all data values are scaled to the interval $[0,1]$ to make them more robust to the ANN-models. Note that the scaler is previously fitted only to the training data in order not to introduce properties of the unseen test data in any way. The models are refitted with a period of 10 days in order to include them in the training set.

Hyperparameter tuning is carried out via random search as proposed by \cite{bergstra2012} combined with subsequent manual fine tuning. Depending on the network, we have to deal with up to 10 different parameters, partly continuously distributed in intervals of several orders of magnitude. All parameters are listed in Table \ref{table:hyperparams}. In that case a pure grid search would not be feasible anymore. The number of epochs in Table \ref{table:hyperparams} refers to the maximum  allowed number of iterations. This limit is not reached by any model because early stopping is used: After not improving (i.e. not reducing the loss) in 50 consecutive runs the model training is terminated. This (like the $L_2$ regularization) is an effective measure to prevent the model from overfitting. Note that, as indicated in Table \ref{table:hyperparams}, in case of using the pinball loss function we treat the upper and lower quantile as hyperparameter. This means that to reach the  95\% quantile, for example, it might be necessary to train the net with regards to the 98\% quantile \cite{jensen2022}.

\begin{table}[h!]
\centering
\begin{tabular}{@{} l | l  l@{}}
\toprule
  Hyperparameter & NN Model & Value Range \\
\hline
 \# Neurons & MLP, LSTM & $4 \dots 128$ \\
 \# Filters & TCN & $4 \dots 128$ \\
 Kernel size & TCN & $3 \dots 7$ \\
 \# Layers & all & $1 \dots 3$ \\
 \# Proceeding days &  all & $1 \dots 12$ \\ 
 Batch size & all & $8 \dots 64$ \\
 \# Epochs & all & $1 \dots 1000$\\
 Learning rate & all & $[1e^{-2} \dots 1e^{-5}]$ \\
 $L_2$ regularization & all & $[1e^{-2} \dots 1e^{-5}]$ \\
Lower quantile & PB & $[0.01 \dots 0.15]$ \\
 Upper quantile & PB & $[0.8 \dots 0.99]$ \\
 Smoothing parameter & QD & $[50 \dots 200]$ \\
 Lagrange multiplier & QD & $[1e^{-2} \dots 1]$ \\
 \bottomrule
\end{tabular}
\caption{All ANN hyperparameters and their range of values}
\label{table:hyperparams}
\end{table}

\subsection{Results}
\label{Sec:Results}

\begin{figure}[ht]
	\centering
		\includegraphics[width=1.0\columnwidth]{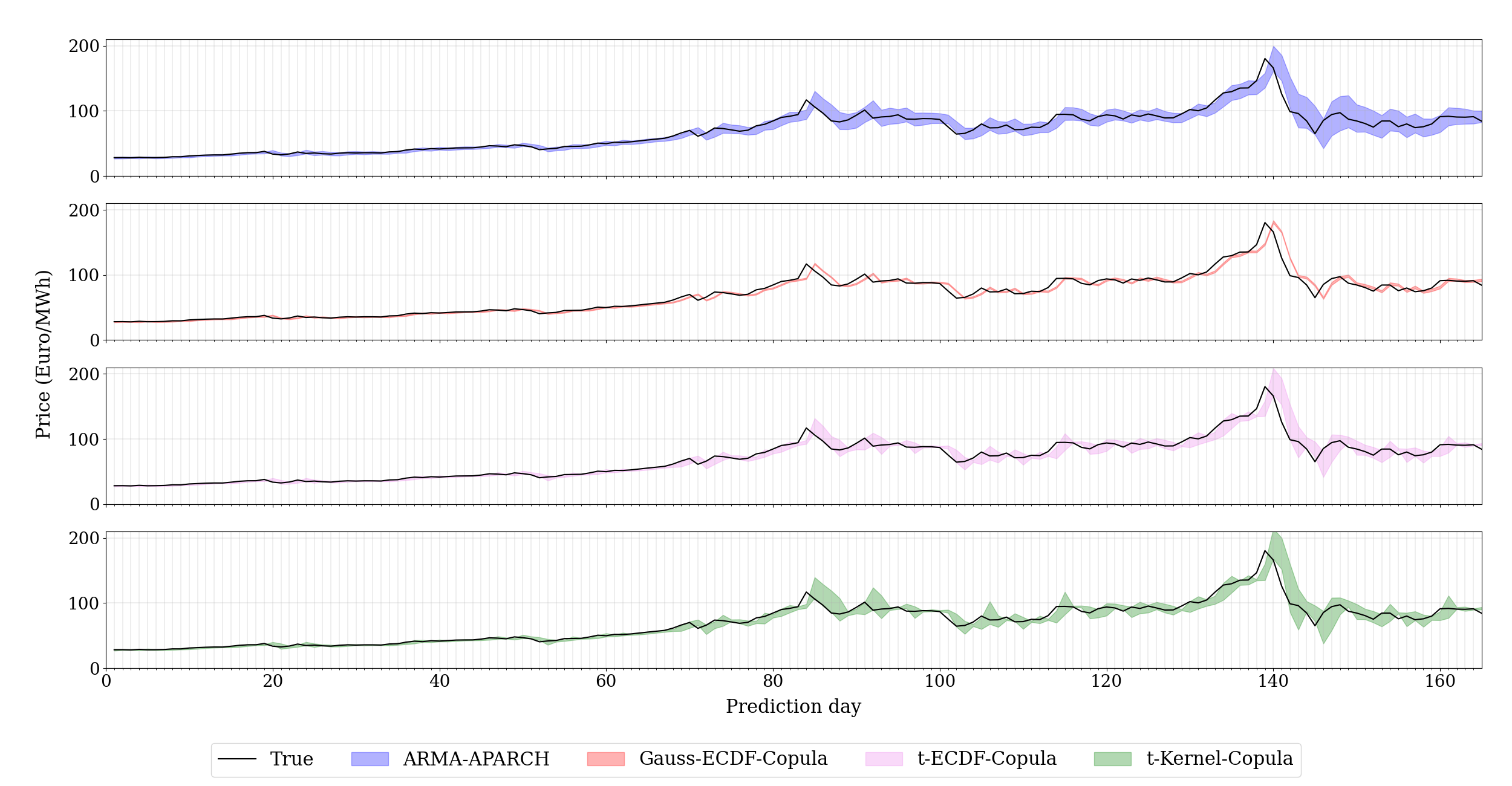}
	\caption{Performance of statistical models during the shock period}
	\label{fig:copula_during}
\end{figure}

\begin{figure}[h!]
	\centering
		\includegraphics[width=1.0\columnwidth]{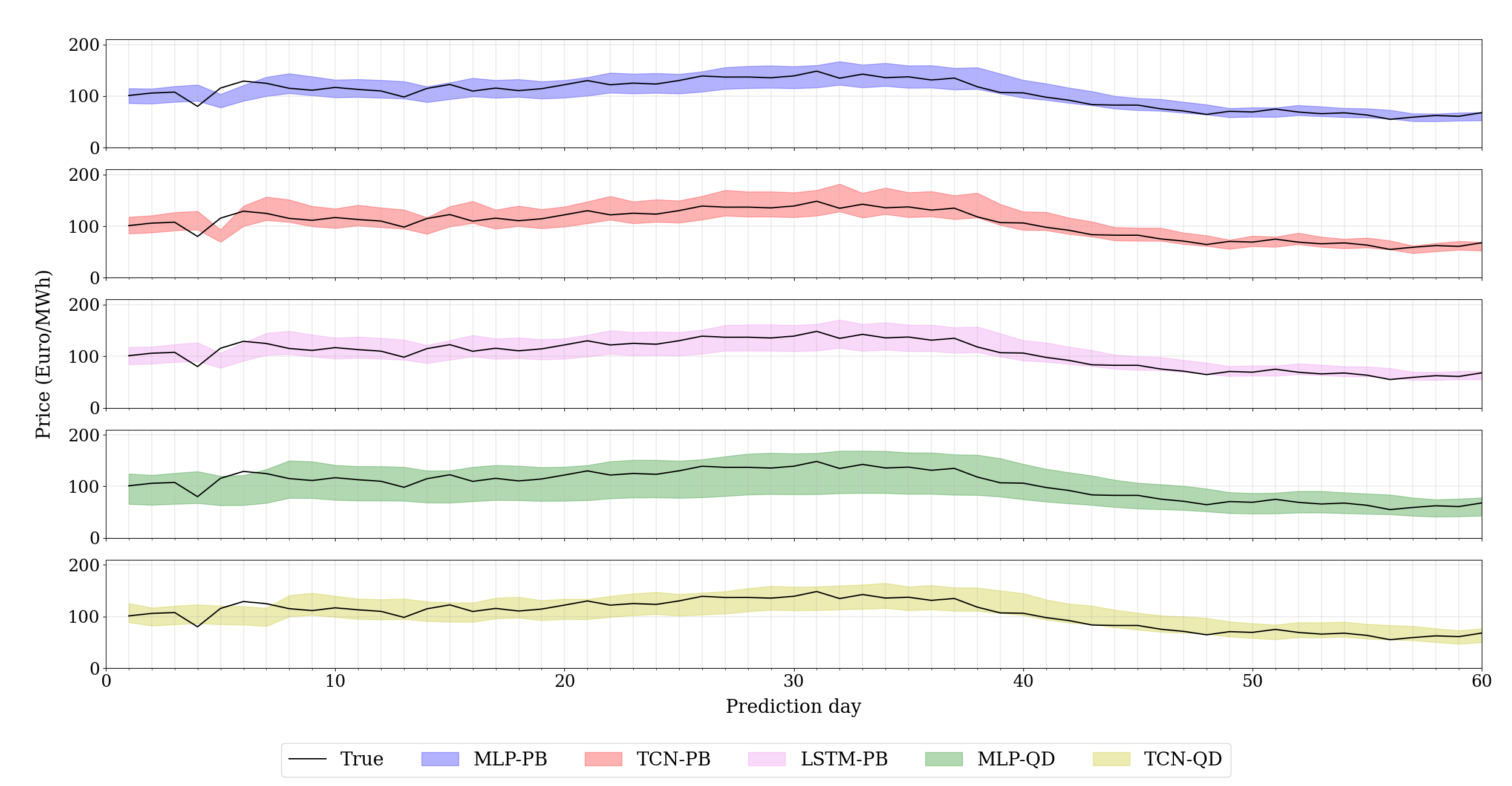}
	\caption{Performance of neural network models in the after-shock period}
	\label{fig:nn_after}
\end{figure}

\begin{figure}[h!]
	\centering
		\includegraphics[width=1.0\columnwidth]{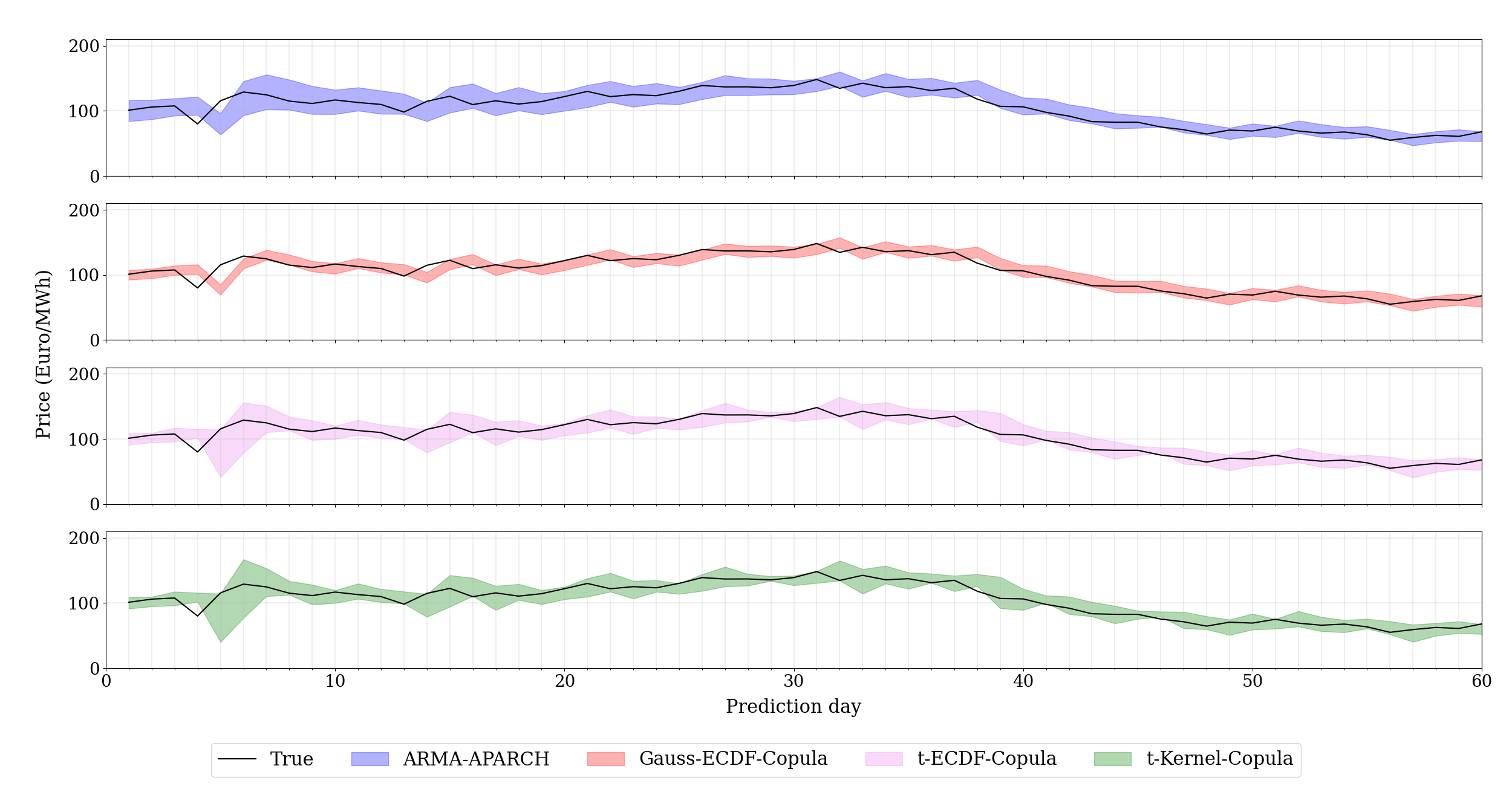}
	\caption{Performance of statistical models in the after-shock period}
	\label{fig:copula_after}
\end{figure}

In order to gain deeper insight into the models' behavior we analyze their performance in two periods separately, namely during what we call the shock period between June 2021 and January 2022 and post-shock period between November 2022 and January 2023. All results are summarized in Tables 6 and 7. In the tables, we report the prediction interval percentage coverage (PICP), the prediction interval average width (PIAW), the interval score (IS) \cite{gneiting2007strictly} and both the $0.05$ and $0.95$ pinball loss $\text{PB}^{0.05}$ loss and $\text{PB}^{0.95}$ loss, respectively. We leave out results for the calm period before summer 2021 as our focus lies on the shock and the time after. In order to better evaluate the models' performance we include graphics that display the prediction intervals over time. Thereby Figures \ref{fig:nn_during} and \ref{fig:copula_during} show the intervals during the shock period. Figures \ref{fig:nn_after} and \ref{fig:copula_after} display the predicted intervals in the after-shock period. Note that in both figures and tables a neural network training based on pinball losses is indicated by PB. The abbreviation QD indicates quality driven losses.

\begin{figure}[ht]
	\centering
		\includegraphics[width=1.0\columnwidth]{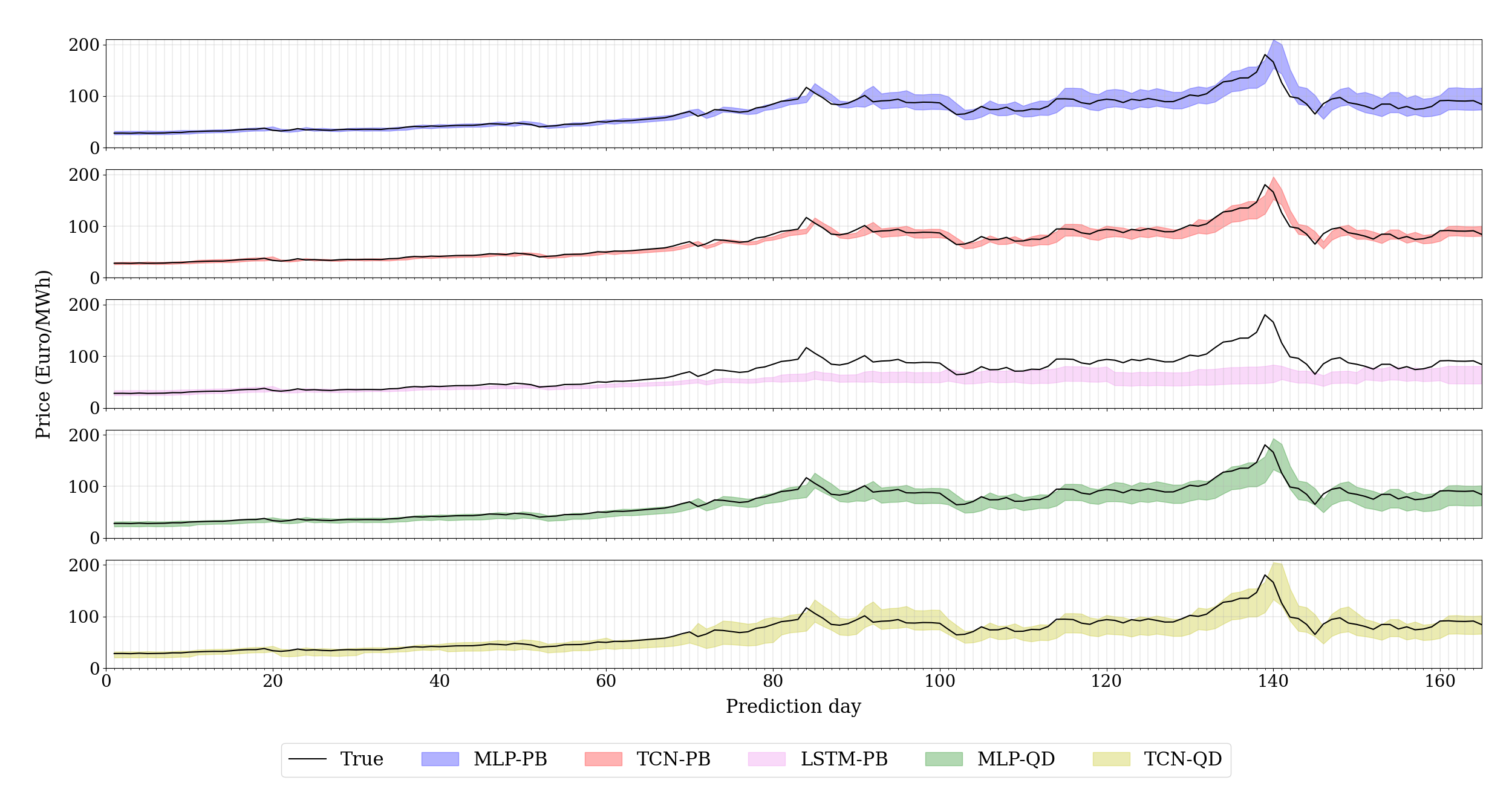}
	\caption{Performance of neural network models during the shock period}
	\label{fig:nn_during}
\end{figure}

One insight from the results is that complexity does not seem to pay off. During the shock period (Table 3), the MLP-QD performs best with regards to the PICP with a value of $0.9073$. 
In the after-shock period (Table 4, the comparably simple benchmark model, ARMA-APARCH, and the MLP-PB  outperform the competitors.
This indicates that rather simple temporal dependence modeling,
can adjust best to changing environments. Examining the pinball losses reveals asymmetry in the performance of most forecasts. During the shock period, the $0.05$ pinball losses of the t copula kernel model, the ARMA-APARCH model, the MLP-PB, the TCN-PB, and the LSTM-PB are smaller than their $0.95$ pinball losses. This indicates that the $0.05$ quantile, i.e. the lower boundary of the PI, is predicted more precisely than the upper boundary of the PI for these models. The opposite holds for the models where the $0.05$ pinball losses are greater than the $0.95$ pinball losses. This insight that incorporating asymmetry into the models might improve forecasting quality in general. 
Regarding the IS during the shock period, the ARMA-APARCH model performs best, and especially better than the MLP-QD. This is indicated by smaller PI widths of the ARMA-APARCH model. It remains to be decided (depending on the application) which criterion is valued higher, i.e. IS or PICP; here, we consider the PICP as the main criterion. Hence, the MLP-QD is considered to be the best model during the shock.  After the shock, the t copula kernel model is performing best with regards to the IS, although its PICP is only $0.8333$. This is again due to the smaller PIAW.

\begin{figure}[htb]
	\centering
		\includegraphics[width=1.0\columnwidth]{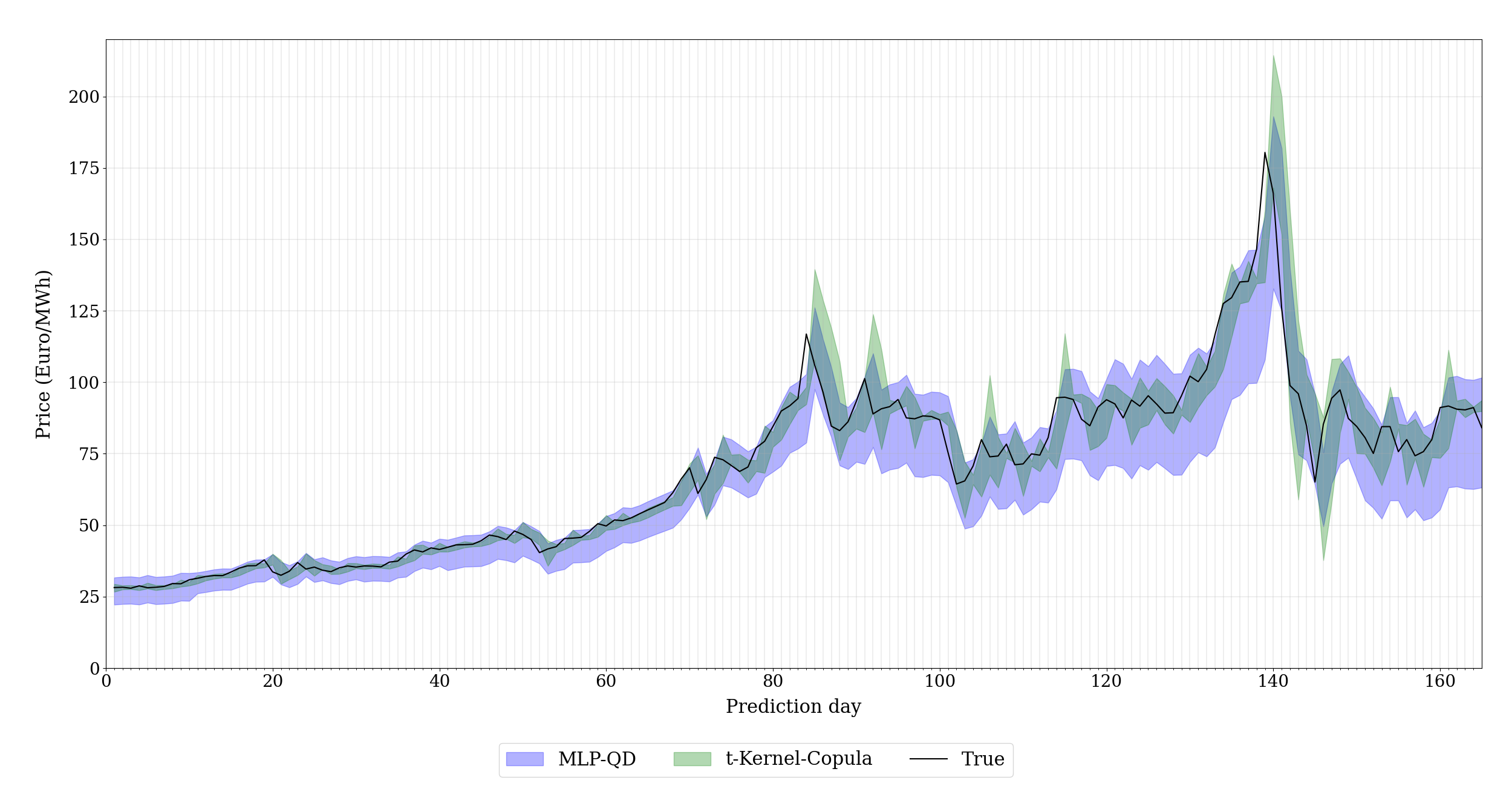}
	\caption{t copula kernel vs. MLP-QD during shock period}
	\label{fig:nn_during}
\end{figure}

\begin{table}[h]
\centering
\begin{tabular}{@{}l | l l l l l@{}} \hline
Model           & PICP     & PIAW      & Interval Score & PB$^{0.05}$-loss & PB$^{0.95}$-loss \\ \hline
t copula kernel   & $0.5818$ & $10.6885$ & $36.7415$  & $0.8535$ & $0.9836$\\
ARMA-APARCH          & $0.8364$ & $14.8305$ & $27.0107$ & $0.6494$ & $0.7012$\\
MLP-PB   & $0.8358$ & $19.1392$ & $33.4463$ & $0.7837$ & $0.8886$\\
MLP-QD & $0.9073$ & $22.1393$ & $30.2696$ & $0.7953$ & $0.7182$\\
TCN-PB   & $0.6297$ & $11.6182$ & $40.2366$ & $0.7130$ & $1.2989$\\
TCN-QD & $0.8824$ & $27.8342$ & $49.4080$ & $1.3286$ & $1.1418$\\
LSTM-PB  & $0.2740$ & $15.7998$ & $262.5863$ & $1.3434$ & $11.7860$\\
\bottomrule
\end{tabular}
\caption{Shock period (2021-06-09 to 2022-01-31) - 0.05 and 0.95 quantile forecast\\ evaluation on last 165 observations}
\label{tab:shockPeriodMLP}
\end{table}

\begin{figure}[htb]
	\centering
		\includegraphics[width=1.0\columnwidth]{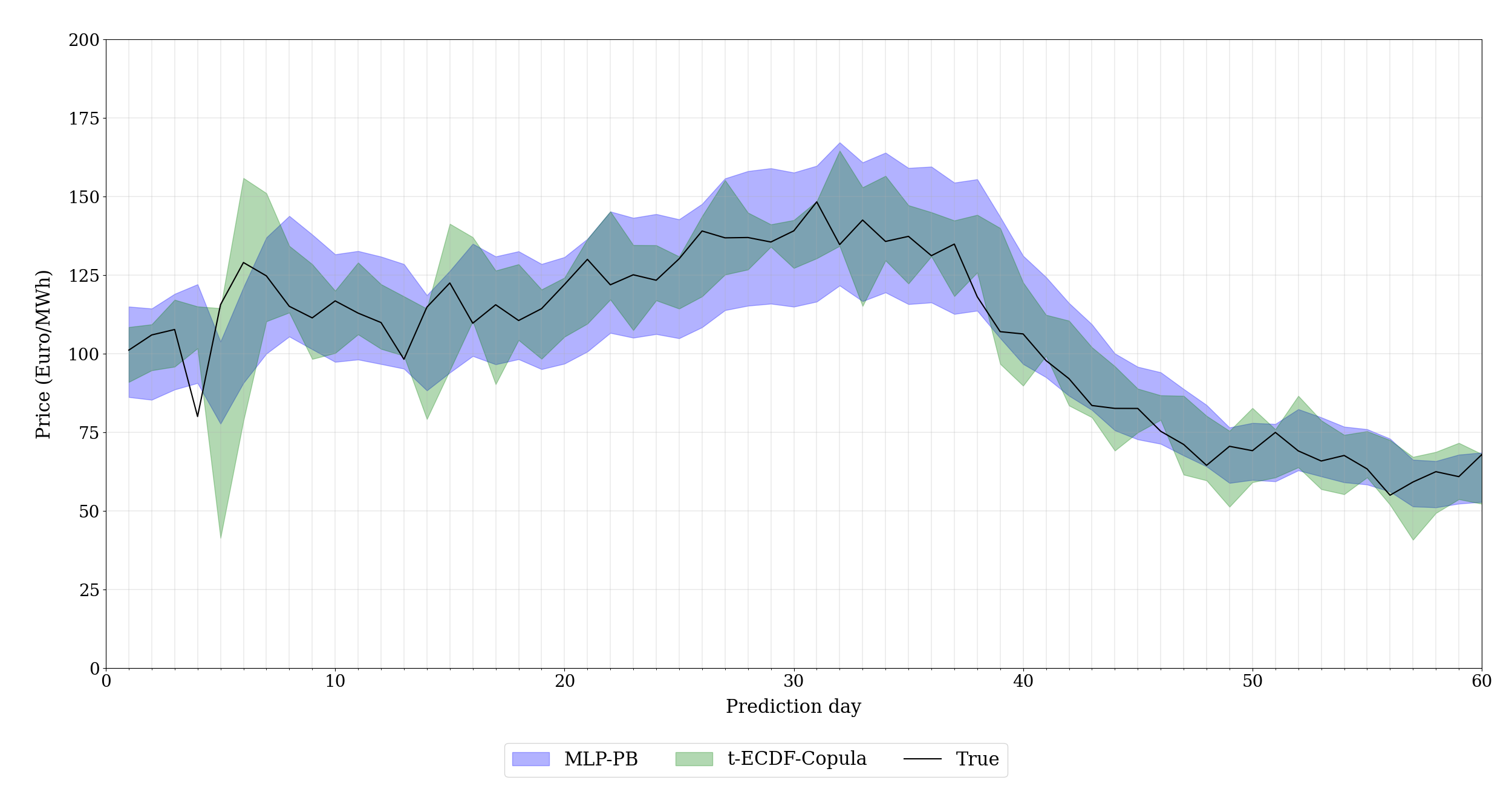}
	\caption{t copula kernel vs. MLP-PB after shock period}
	\label{fig:nn_during}
\end{figure}

\begin{table}[h]
\centering
\label{tab:AftershockPeriodMLP}
\begin{tabular}{@{}l | l l l l l@{}} \hline
Model                & PICP      & PIAW       & Interval Score & PB$^{0.05}$-loss & PB$^{0.95}$-loss\\ \hline
t copula kernel   &  $0.8333$    & $25.0831$  & $38.5717$ & $1.2361$ & $0.6925$\\
ARMA-APARCH          &  $0.8833$    & $27.0702$  & $42.3472$ & $1.0241$ & $1.0933$ \\
MLP-PB   &  $0.8850$ & $31.3743$  & $43.7791$ & $0.9965$ & $1.1925$\\
MLP-QD &  $0.9783$ & $61.5754$  & $64.5194$ & $1.8516$ & $1.3744$\\
TCN-PB   &  $0.8950$ & $35.0266$  & $51.2600$ & $0.9398$ & $1.6202$\\
TCN-QD &  $0.8583$ & $37.6244$  & $52.3602$ & $1.2393$ & $1.3787$\\
LSTM-PB  &  $0.8950$ & $36.2149$ & $48.1500$ & $1.1493$ & $1.2582$\\
\bottomrule
\end{tabular}
\caption{Calm period after the shock (2022-10-26 - 2023-01-20) - 0.05 and 0.95 quantile forecast\\ evaluation on last 60 observations}
\end{table}

During the shock period, the predicted interval widths of all models --except for MLP-QD -- are too small as they are calibrated mainly on the pre-shock time series (see Figures \ref{fig:nn_during} and \ref{fig:copula_during}). In contrast to this, 
forecasting in the after-shock period using updated data (which has more recent values, from the shock period in particular) results in interval forecasts being too big (see Figures \ref{fig:nn_after} and \ref{fig:copula_after}). Note that this result holds in moving window as well as in expanding window forecasting studies. The models pick up the recent behavior of the time series and hence considerably overestimate variances in the after-shock period. There are two underlying mechanisms causing this behavior. First, including data from the shock period in the estimation results in the parameters reflecting the extreme behavior during the shock period. Second, even when parameters from before the shock period are used in the models and forecasts are conducted for the after-shock period, there could still be an effect from the shock in persistent time series models. Through moving average terms, the effect from the shock is transmitted to all future forecasts (of course with decreasing amplitude). This effect is only possible in the ARMA-APARCH model and the LSTM.

Interestingly, the LSTM model, which is highly regarded in time series forecasting, performs worst during the volatile period. As displayed in Figure \ref{fig:nn_during} it is completely off track in the shock period; however, its interval width is corrected in the after-shock period (Figure \ref{fig:nn_after}). 
To improve the poor performance of the LSTM network, several things are tested: To enhance the input data, we create more features by calculating the 1st - 4th moments of the input values based on the first four (orthogonal) Chebyshev polynomials. This is proposed by \cite{ormaniec2022} as they also have to face a lack of training data (in that case 7500 data points) for their LSTM model. Alternatively adaptions to the LSTM architecture are tested: The depth and the width of the networks is increased and weight regularization based on the L2-norm  are added. Finally an additional batch normalization layer is included after the last hidden layer. As a consequence results are improved to some extent, but the network still performs very unstable as training is presumably often stuck in a local minimum. In order to investigate the influence of historical data, we roughly interpolate the peak from July 2022 from the test data and insert it into the training data. It shows that the network indeed reacts sensible to the test data; now the LSTM performance is similar to the other network models.
The t-copula model is not able to capture conditional heteroskedasticity entirely during the shock period. This may be caused by the degree of freedom of the t-copula being too high. We also conducted the forecasting study using the empirical distribution function as marginal distribution, however, there was no improvement. Experimentation with the Gaussian copula yielded the expected result: The Gaussian copula is not able to capture conditional heteroskedasticity and is not able to adapt to higher conditional variances. Hence, the PIs are too small. The PIs generated by these models are also displayed in Figure~\ref{fig:copula_during} and \ref{fig:copula_after}, respectively.

\section{Discussion}
\label{subsec:discuss}

The results from Section \ref{Sec:Results} indicate that in troubled times complex concepts such as an LSTM model have to be discarded or need to be improved respectively. Thereby the weak performance of the LSTM can be explained by the absence of mid to long term patterns due to the structural break and a high level of heteroskedasticity especially in the latest prices. However, shortening the calibration set is also no alternative as in this case we have an imbalance between a large number of parameters and a small training set. To cite \cite{hill1996}, in this case "the model tends to memorize the data rather then generalize from it". ARIMA, on the other side, relies only on a few fundamental parameters and can be calibrated using a comparably short data history. Hence the model is able to adapt quicker to structural changes. This also holds true for the GARCH model (and versions of it). It is actually designed to incorporate volatility clusters so reacts quickly to changes in the volatility structure. However, there are still options to improve forecasting quality. Statistical models are trained to minimize log-likelihood, not pinball loss. Training with respect to the pinball loss could enhance the performance. It would also be possible to train with respect to the interval score (see \cite{gneiting2007strictly}).

Regarding the chosen neural networks, one could test whether more features in form of related time series might improve the forecasting quality. This can be comparably easy done for neural networks but in the copula case complexity increases considerably. Besides, additional data require additional data handling efforts (missing values, skewed values, more interdependencies). As an indication for future research it was tested whether electricity prices might increase interval forecasting accuracy, but the effect can be neglected. It still needs to be verified which external data and at what granularity (hourly, daily) helps to improve forecasts. Ensemble learning such as proposed by \cite{zhang2007neural} or \cite{gunturi2021ensemble} is another option, and future research needs to show whether a combination of different methods helps to improve the forecasting quality. However, first attempts in the context of this case study were not promising. 

Eventually, one has to discuss whether constructing PIs based on QR algorithms is the right approach. Such PIs are often not valid as they don't fit the designated confidence level. Here conformal prediction \cite{angelopoulos2021} offers a solution, as it constructs valid PIs but with only slightly varying length. A combination of both methods, which comes along with a considerable increase of complexity, might lead to narrower yet valid PIs. The resulting method is called conformal quantile regression (CQR)  \cite{jensen2022} \cite{romano2019conformalized}. Another alternative is mean-variance estimation (MVE) for PI construction as proposed by by \cite{nix1994}. The method uses a neural network to estimate the characteristics of the conditional target distribution. Additive Gaussian noise with time-varying variance is assumed. However, this method underestimates the variance of data, leading to a low empirical coverage probability, as discussed in \cite{ding2003} and \cite{dybowski2000}. Bayesian theory also offers  a range of methods, but they impose comparably hard expectations/assumptions on the data. Besides, existing Bayesian techniques lack scalability to large dataset and network sizes \cite{hernandez15}.
Moreover, Bayesian methods are computationally demanding in the development stage. It requires calculation of the Hessian matrix  (usually with Levenberg–Marquardt optimization) which is time consuming and cumbersome for large NNs and data sets \cite{khosravi2011b}.
Despite the strength of the supporting theories, the method suffers from massive computational
burden \cite{khosravi2011}.

\section{Conclusion}
\label{secConclusion}

Reliable forecasts of gas prices are needed in various applications and especially in risk management and energy trading. This article analyzes both classical statistical methods and neural network-based models and compares their performance with regards to a rather simple autoregressive approach. Thereby, interval forecasts for Dutch Front Month gas prices are considered. It shows that neural networks perform reasonably well in the period before the price (and volatility) increase in Summer 2022, but are clearly outperformed during and after the shock phase. The LSTM model, an approach widely used for forecasting, shows to be the worst model and has to be discarded. It is assumed that especially the long-memory component struggles to include the structural break in Summer 2022. Future research is needed to see whether model training can be improved. Eventually, it shows that the simplest model, i.e. an ARMA-APARCH model, performs best.

As summarized in the discussion section there are few possibilities to improve forecasting quality of both copula- and neural network-based models that still need to be tested. It also needs to be verified whether more complex methods such as conformalized quantile regression are able to improve  the results. Future research is needed.

\section{References}
\bibliographystyle{plain}
\bibliography{myBib}

\end{document}